\documentclass[journal]{IEEEtran}
\usepackage{gensymb}
\usepackage{amsmath,amssymb}
\usepackage{psfrag}
\usepackage{epsfig}
\usepackage{cite}
\usepackage{graphics}
\usepackage{subfigure}
\usepackage{multirow}
\usepackage{threeparttable}
\usepackage{booktabs}
\usepackage{mathrsfs}
\usepackage{algorithmicx,algorithm}
\usepackage{hyperref}
\usepackage{algorithmicx}
\usepackage{algpseudocode}

\begin{document}
\title{Egret Swarm Optimization Algorithm: An Evolutionary Computation Approach for Model Free Optimization}

\author{Zuyan Chen, Adam Francis, Shuai Li, Bolin Liao, Dunhui Xiao

	\thanks{Zuyan Chen, Adam Francis, and Shuai Li are with School of Engineering, Faculty of Science and Engineering, Swansea University, Swansea, SA1 8EN, UK. 
		E-mails: zuyan\_chen@outlook.com (Zuyan, Chen),
		a.g.francis.917537@swansea.ac.uk (Adam Francis),
		shuai.li@swansea.ac.uk (Shuai Li); Bolin Liao is with School of Information Science and Engineering, Jishou University, Jishou, 416000, Hunan, China; Dunhui Xiao is with School of Mathematical Sciences,Tongji University, Shanghai, P.R. China,200092; E-mail: xiaodunhui@tongji.edu.cn.
	} 
}

\markboth{}%
{Shell \MakeLowercase{\textit{et al.}}: Bare Demo of IEEEtran.cls
	for Journals} \maketitle

\begin{abstract}
	A novel meta-heuristic algorithm, Egret Swarm Optimization Algorithm (ESOA), is proposed in this paper, which is inspired by two egret species' (Great Egret and Snowy Egret) hunting behavior. ESOA consists of three primary components: Sit-And-Wait Strategy, Aggressive Strategy as well as Discriminant Conditions. The performance of ESOA on 36 benchmark functions as well as 2 engineering problems are compared with Particle Swarm Optimization (PSO), Genetic Algorithm (GA), Differential Evolution (DE), Grey Wolf Optimizer (GWO), and Harris Hawks Optimization (HHO). The result proves the superior effectiveness and robustness of ESOA. The source code used in this work can be retrieved from \href{https://github.com/Knightsll/Egret\_Swarm\_Optimization_Algorithm}{https://github.com/Knightsll/Egret\_Swarm\_Optimization\_Algorithm}; \href{https://ww2.mathworks.cn/matlabcentral/fileexchange/115595-egret-swarm-optimization-algorithm-esoa}{https://ww2.mathworks.cn/matlabcentral/fileexchange/115595-egret-swarm-optimization-algorithm-esoa}. 
\end{abstract}

\begin{IEEEkeywords}
	Egret Swarm Optimization Algorithm, Meta-Heuristic Algorithm, Constrained Optimization, Swarm Intelligence.
\end{IEEEkeywords}

\section{Introduction}
\begin{figure*}\centering
	
	\includegraphics[scale=0.17]{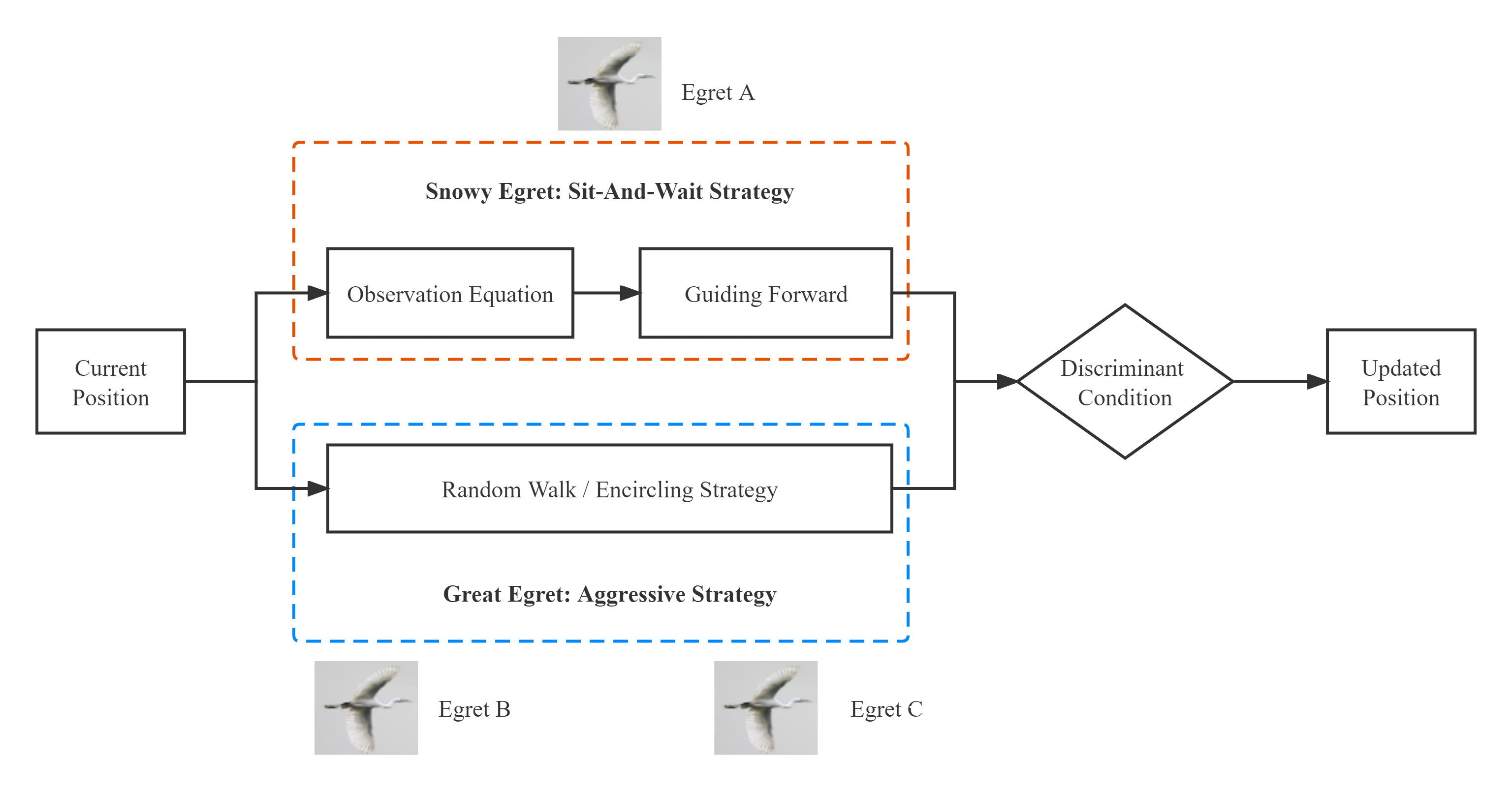}
	
	\caption{The main framework of ESOA.}\label{fig.1}
\end{figure*}
General engineering applications involving manipulator control, path planning, and fault diagnosis can be described as optimization problems. Since these problems are almost non-convex, conventional gradient approaches are difficult to apply and frequently result in local optima \cite{ref.1}. In recent years, due to the global optimal, gradient-free, and high-efficiency properties, meta-heuristics algorithms have been utilized to solve practical engineering problems. 

Meta-heuristic algorithms imitate natural phenomena through simulating animal and other environmental behaviours. These algorithms are broadly inspired by three concepts: species evolution, biological behavior, and physical principles \cite{ref.2}. Evolution-based algorithms generate various status solution spaces via mimicking species evolution's diversity, cross-mutation, and Survival of the fittest. The Darwinian evolution-inspired Genetic Algorithm (GA) has remarkable global search capabilities and has been applied in a variety of disciplines \cite{ref.3}. Comparable to GA, Differential Evolution (DE) has also demonstrated strong adaptiveness in a variety of optimizations \cite{ref.4}. Physics-based approaches apply the physical law as a way to achieve an optimal solution. The Simulated Annealing (SA) \cite{ref.5}, Gravitational Search Algorithm (GSA) \cite{ref.6}, Black Hole Algorithm (BH) \cite{ref.7}, and Multi-Verse Optimizer (MVO) \cite{ref.8} are the common methods of this type.

The biological behavior-based algorithms mimic organisms' behavior such as predation, pathfinding, growth, and aggregation in order to solve numerical optimization problems. Particle Swarm Optimation (PSO), a swarm intelligence algorithm inspired by bird flocking behavior, is the typical representative \cite{ref.9}. PSO leverages the exchange of information among individuals in a population to develop the entire population's motion from disorder to order in the problem-solving space, resulting in an optimal solution. Moreover, \cite{ref.10} was inspired by the behaviour of ant colonies, proposing the Ant Colony Optimization (ACO) algorithm. ACO indicates a feasible solution to the optimization problem in terms of the ant pathways, with shorter paths depositing more pheromones. The concentration of pheromone collecting on the shorter pathways steadily rises over time, and the entire ant colony will focus on the optimal path due to the influence of positive feedback. 

With the widespread application of PSO and ACO in areas such as robot control, route planning, artificial intelligence, and combinatorial optimization, heuristic algorithms have spawned a plethora of excellent research. \cite{ref.11} presented Grey Wolf Optimizer (GWO) based on the hierarchy and hunting mechanism of grey wolves. \cite{ref.12} applied GWO to restructure a maximum power extraction model for a photovoltaic system under a partial shading situation. GWO also was utilized in nonlinear servo systems in order to tune the Takagi-Sugeno parameters of the proportional-integral-fuzzy controller \cite{ref.13}. \cite{ref.14} introduced the sparrow search algorithm (SSA) inspired by sparrows' collective behavior, foraging, and anti-predation activities. \cite{ref.15} suggested a novel incremental generation model based on a chaotic sparrow search algorithm to handle large-scale data regression and classification challenges. An integrated optimization model for dynamic reconfiguration of active distribution networks was built with multi-objective SSA by \cite{ref.16}. \cite{ref.17} constructed a tiny but efficient meta-heuristic algorithm named Beetle Antennae Search Algorithm (BAS), through modeling the beetle's predatory behavior. \cite{ref.18} integrated BAS with a recurrent neural network to create a novel robot control framework for redundant robotic manipulator trajectory planning and obstacle avoidance. \cite{ref.19} applied BAS to optimize the initial parameters of the convolutional neural network for medical imaging diagnosis, demonstrating excellent accuracy and the short period of tuning time.

Although meta-heuristics algorithms have achieved marvelous performance in various engineering applications, as Wolpert analyses in \cite{ref.20}, there is no near-perfect method that can deal with all optimization problems. To put it another way, if an algorithm is appropriate for one class of optimization problems, it may not be acceptable for another. Alternatively, the search efficiency of an algorithm is inversely related to its computational complexity, and a certain amount of computational consumption needs to be sacrificed in order to enhance search efficiency. The No Free Lunch (NFL) theorem has assured that the field has flourished, with new structures and frameworks for meta-heuristics algorithms constantly emerging.

This paper proposes a novel meta-heuristic algorithm (Egret Swarm Optimization Algorithm, ESOA) to examine how to improve the balance between the algorithm's exploration and exploitation. The rest of the paper is structed as follows:
Section 2 depicts the observation of egret migration behavior as well as the development of the ESOA framework and mathematical model. The comparison of performance and efficiency in CEC1997 and CEC2017 between ESOA and other algorithms is demonstrated in Section 3. The result and convergence of two engineering optimization problems utilizing ESOA are discussed in Section 4. Section 5 represents the conclusion of this paper as well as outlines further work.

\section{Egret Swarm Optimization Algorithm}

This section reveals the inspiration of ESOA. Then, the mathematical model of the proposed method is discussed. 

\subsection{Inspiration}
The egret is the collective term for four bird species: the Great Egret, the Middle Egret, the Little Egret, and the Yellow-billed Egret, all of which are known for their magnificent white plumage. The majority of egrets inhabit coastal islands, coasts, estuaries, and rivers, as well as lakes, ponds, streams, rice paddies, and marshes near their shores. Egrets are usually observed in pairs, or in small groups, however vast flocks of tens or hundreds can also be spotted \cite{ref.21, ref.22, ref.23}. Maccarone observed that Great Egret fly at an average speed of 9.2m/s and balance their movements and energy expenditure whilst hunting \cite{ref.24}. Due to the high consumption of energy when flying, the decision to prey typically necessitates a thorough inspection of the trajectory to guarantee that it may obtain more energy through the location of food than what would be expended through flight. Compared to Great Egrets, Snowy Egrets tend to sample more sites, and they will observe and select the location where other birds have already discovered food \cite{ref.25}.
Snowy Egrets often adopt a sit-and-wait strategy, a scheme that involves observing the behavior of prey for a period of time and then anticipating its next move in order to hunt with the least energy expenditure \cite{ref.28}. Maccarone indicated in \cite{ref.27} that not only do Snowy Egrets which apply the strategy consume less energy, but they are also 50\% more efficient at catching prey than other egrets. However, although the Great Egret adopt a higher exertion strategy to pursue its prey aggressively, they are capable of capturing more plentiful prey since it is rare for larger prey to travel through an identical place multiple times \cite{ref.26}. Overall, Great Egrets with aggressive strategies exchange high energy consumption for potentially greater returns, whereas Snowy Egrets with a sit-and-wait approach, lower energy expenditure is able to be swapped for more reliable profits \cite{ref.27}.

\subsection{Mathematical Model And Algorithm}

Inspired by the Snowy Egret's Sit-And-Wait strategy and the Great Egret's Aggressive strategy, ESOA has combined the advantages of both strategies and constructed a corresponding mathematical model to quantify the behaviors. As shown in the Fig. \ref{fig.1}, ESOA is a parallel algorithm with three essential components: the Sit-And-Wait Strategy, the Aggressive Strategy, and the Discriminant Condition. There are three Egrets in one Egret squad, Egret A applied guiding forward while Egret B and Egret C adopt random walk and encircling mechanisms respectively. Each part is detailed below.

\subsubsection{Sit-And-Wait Strategy}
Observation Equation: Assuming that the position of the i-th egret squad is $\mathbf{x}_i \in \mathbb{R}^n$, $n$ means the dimension of problem, $A(\ast)$ presents Snowy Egret's estimate approach of the possible presence of prey in its own current location. $\hat{y}$ means the estimate value of prey in current location. 

\begin{equation}
	\hat{y}_i = A(\mathbf{x}_i),
\end{equation}
then the estimate method could be parameterized as,

\begin{equation}
	\hat{y}_i = \mathbf{w}_i \cdot \mathbf{x}_i,
\end{equation}
where the $\mathbf{w}_i \in \mathbb{R}^n$ is the weight of estimate method. The error $e_i$ could be described as below,

\begin{equation}
	e_i = \left \| \hat{y}_i - y_i \right \|^2/2 \label{eq.3}.
\end{equation}
Meanwhile, $\hat{\mathbf{g}}_i \in \mathbb{R}^n$, the practical gradient of $\mathbf{\omega}_i$, could be retrieved via taking the partial derivative of $\mathbf{\omega}_i$ for the error Eq. (\ref{eq.3}), and its direction is $\hat{\mathbf{d}}_i$. 

\begin{equation}
	\begin{aligned}
		\hat{\mathbf{g}}_i &= 
		\frac{\partial \hat{e}_i}{\partial \mathbf{w}_i} \\
		&= \frac{\partial \left \| \hat{y}_i - y_i \right \|^2/2}{\partial \mathbf{w}_i} \\
		&= (\hat{y}_i - y_i) \cdot \mathbf{x}_i, \\
		\hat{\mathbf{d}}_i &= \hat{\mathbf{g}}_i/  \left | \hat{\mathbf{g}}_i \right |.
	\end{aligned}\label{eq.4}
\end{equation}

\begin{figure}[tb]\centering
	
	\includegraphics[scale=0.2]{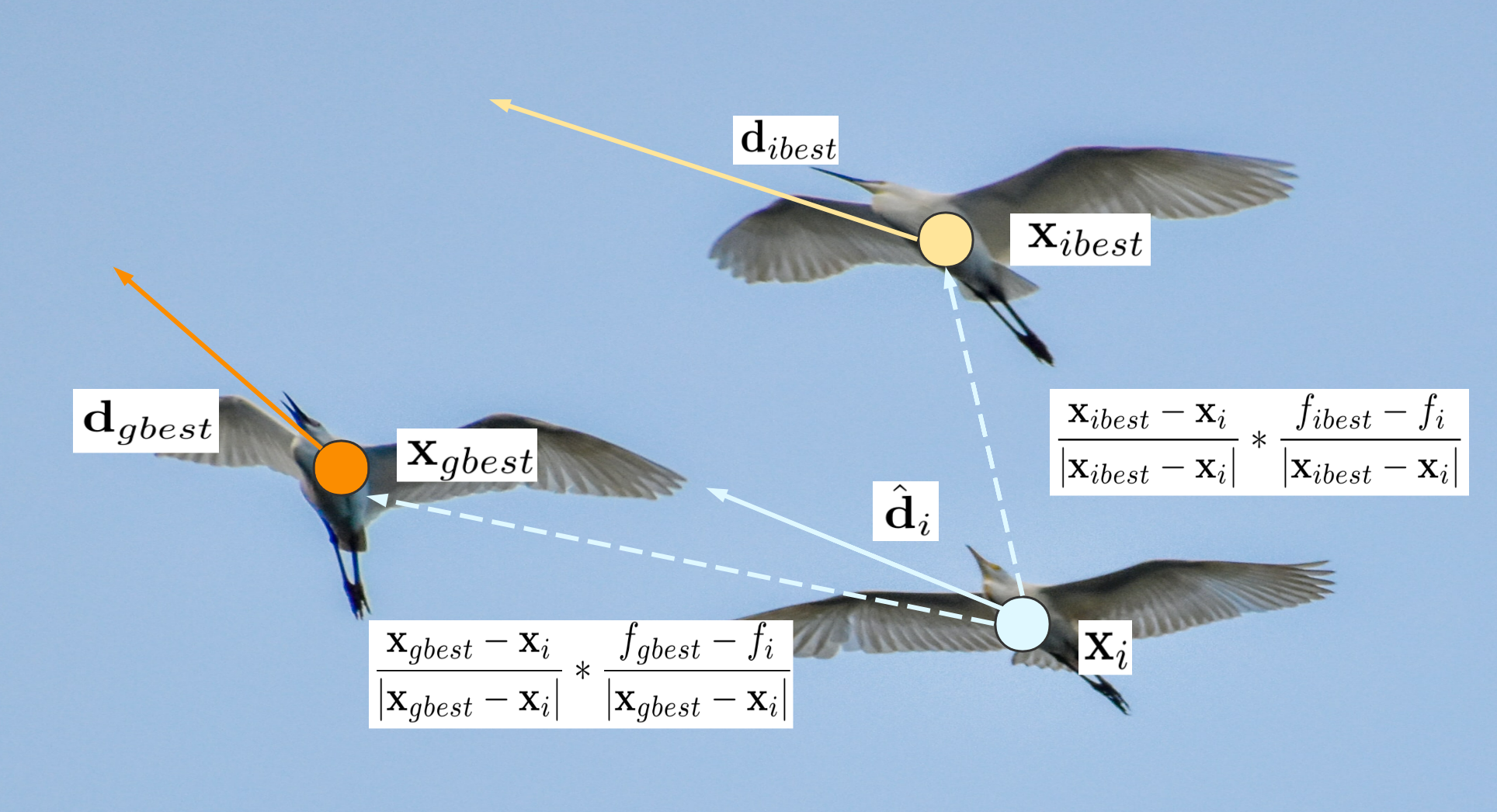}
	
	\caption{Following behaviour of egret Swarms as a effective way for gradient estimation.}\label{fig.2}
\end{figure}

Fig.\ref{fig.2} demonstrates egret's following behavior, where egrets refer to better egrets during preying, drawing on their experience of estimating prey behavior and incorporating their own thoughts. $\mathbf{d}_{h,i} \in \mathbb{R}^n$ means the directional correction of the best location of the squad while $\mathbf{d}_{g,i} \in \mathbb{R}^n$ presents the the directional correction of the best location of all squad.

\begin{equation}
	\mathbf{d}_{h,i} = \frac{\mathbf{x}_{ibest} - \mathbf{x}_i}{\left | \mathbf{x}_{ibest} - \mathbf{x}_i \right | }
	\cdot\frac{f_{ibest} - f_i}{\left | \mathbf{x}_{ibest} - \mathbf{x}_i \right | } + \mathbf{d}_{ibest}.\label{eq.5}
\end{equation}

\begin{equation}
	\mathbf{d}_{g,i} = \frac{\mathbf{x}_{gbest} - \mathbf{x}_i}{\left | \mathbf{x}_{gbest} - \mathbf{x}_i \right | }
	\cdot\frac{f_{gbest} - f_i}{\left | \mathbf{x}_{gbest} - \mathbf{x}_i \right | } + \mathbf{d}_{gbest}.\label{eq.6}
\end{equation}
The integrated gradient $\mathbf{g}_i \in \mathbb{R}^n$ could be represented as below, and $r_h \in [0, 0.5)$, $r_g \in [0,0.5)$:

\begin{equation}
	\mathbf{g}_i = (1 - r_h - r_g) \cdot \hat{\mathbf{d}}_i+r_h \cdot \mathbf{d}_{h, i} + r_g \cdot \mathbf{d}_{g,i}.\label{eq.7}
\end{equation}
An adaptive weight update method is applied here \cite{ref.29}, $\beta_1$ is 0.9 and $\beta_2$ is 0.99:

\begin{equation}
	\begin{aligned}
		\mathbf{m}_{i} &= \beta_1\cdot\mathbf{m}_{i} + (1 - \beta_1)\cdot\mathbf{g}_{i},\\
		\mathbf{v}_{i} &= \beta_1\cdot\mathbf{v}_{i} + (1 - \beta_1)\cdot\mathbf{g}^2_{i},\\
		\mathbf{w}_{i} &= \mathbf{w}_{i} - \mathbf{m}_{i}/\sqrt{\mathbf{v}_{i}}.
	\end{aligned}\label{eq.8}
\end{equation}
According to Egret A's judgement of the current situation, the next sampling location $\mathbf{x}_{a,i}$ could be described as,

\begin{equation}
	\mathbf{x}_{a,i} = \mathbf{x}_i + \exp({-t/(0.1\cdot t_{max})}) \cdot 0.1 \cdot hop \cdot \mathbf{g}_i,\label{eq.9}
\end{equation}

\begin{equation}
	y_{a,i} = f(\mathbf{x}_{a,i}),\label{eq.10}
\end{equation}
where $t$ and $t_max$ presents the current iteration times and the maximum iteration times, while $hop$ means the gap between the low bound with the up bound of solution space. $y_{a,i}$ is the fitness of $\mathbf{x}_{a,i}$.

\subsubsection{Aggressive Strategy}
Egret B tends to randomly search prey and its behavior could be depicted as below,

\begin{equation}
	\mathbf{x}_{b,i} = \mathbf{x}_{i} + \tan{(\mathbf{r}_{b,i})}\cdot hop/(1+t),\label{eq.11}
\end{equation}

\begin{equation}
	y_{b,i} = f(\mathbf{x}_{b,i}),\label{eq.12}
\end{equation}
where $\mathbf{r}_{b,i}$ is a random number in $(-\pi/2, \pi/2)$, $\mathbf{x}_{b,i}$ is Egret B's expected next location and $y_{b,i}$ means the fitness.

Egret C prefer to pursue the prey aggressively, so the encircling mechanism is utlized as the update method of its position:
\begin{equation}
	\begin{aligned}
		\mathbf{D}_{h} &= \mathbf{x}_{ibest} -  \mathbf{x}_{i},\\
		\mathbf{D}_g &= \mathbf{x}_{gbest} -  \mathbf{x}_{i},\\
		\mathbf{x}_{c,i} &= (1-\mathbf{r}_i-\mathbf{r}_g)\cdot\mathbf{x}_{i} +\mathbf{r}_h\cdot\mathbf{D}_h + \mathbf{r}_g\cdot\mathbf{D}_g,
	\end{aligned}\label{eq.13}
\end{equation}

\begin{equation}
	y_{b,i} = f(\mathbf{x}_{b,i}). \label{eq.14}
\end{equation}
$\mathbf{D}_h$ is the gap matrix between current location and the best position of this egret squad while $\mathbf{D}_{g}$ compares with the best location of all egret squads. $\mathbf{x}_{c,i}$ means the expected location of Egret C. $r_h$ and $r_g$ are random numbers in $[0, 0.5)$.

\subsubsection{Discriminant Condition}

After each member of the Egret squad has decided on its plan, the squad would select the optimal option and takes the action together. $\mathbf{x}_{s,i}$ is the solution matrix of i-th Egret squad:

\begin{equation}
	\mathbf{x}_{s,i} = \left [ \mathbf{x}_{a,i} \qquad \mathbf{x}_{b,i} \qquad \mathbf{x}_{c,i} \right ], \label{eq.15}
\end{equation}

\begin{equation}
	\mathbf{y}_{s,i} = \left [ y_{a,i} \qquad y_{b,i} \qquad y_{c,i} \right ], \label{eq.16}
\end{equation}

\begin{equation}
	c_{i} = argmin(\mathbf{y}_{s,i}),\label{eq.17}
\end{equation}

\begin{equation}
	\mathbf{x}_{i} = \begin{cases}
		\mathbf{x}_{s,i}|_{c_{i}} \quad if \quad \mathbf{y}_{s,i}|_{c_{i}} < y_{i} \quad or \quad r<0.3, \\
		\mathbf{x}_{i}\qquad\qquad\quad else
	\end{cases}.\label{eq.18}
\end{equation}
If the minimal value of $\mathbf{y}_{s,i}$ is better than current fitness $y_i$, the Egret squad would accept the choice. Or if the random number $r\in(0,1)$ is less than 0.3, which means there is 30\% possibility to accept a worse plan. 

\subsection{Pseudo Code}

Based on the discussion above, the pseudo-code of ESOA is constructed as Algorithm \ref{pseudo}, which contains two main functions to retrieve the Egret squad's expected position matrix and a discriminant condition to choose a better scheme. ESOA requires an initial matrix $\mathbf{x}^0\in\mathbb{R}^{P \text{x} N}$ of the $P$ size Egret Swarm position as input, while it returns the optimal position $\mathbf{x}_{best}$ and fitness $y_{best}$.

\renewcommand{\algorithmicrequire}{\textbf{Input:}}  
\renewcommand{\algorithmicensure}{\textbf{Output:}} 

\begin{algorithm}[h]
	\caption{Egret Swarm Optimization Algorithm}\label{pseudo} 
	\begin{algorithmic}[1]
		\Require
		$\mathbf{x}^0$: the $P$ size Egret Swarm position $\in\mathbb{R}^{P \text{x} N}$;
		\Ensure
		$\mathbf{x}_{best}$: Optimal or approximate optimal solution;
		$y_{best}$: Optimal or approximate optimal fitness;
		
		\Function {Sitandwait}{$\mathbf{x}$}
		\State Update the integrated gradient $\mathbf{g}$ via Eq. (\ref{eq.4} to \ref{eq.7})
		\State Update the weight of observation method $\omega$ by Eq. (\ref{eq.8}) 
		\State Get the expected position $\mathbf{x}_{a}$ of Egret A by Eq.(\ref{eq.9})
		\State \Return{$\mathbf{x}_{a}$}
		\EndFunction

		\Function {Aggressive}{$\mathbf{x}$}
		\State Get the expected position $\mathbf{x}_{b}$ of Egret B by Eq. (\ref{eq.11})
		\State Get the expected position $\mathbf{x}_{c}$ of Egret C by Eq. (\ref{eq.13}) 
		\State \Return{$\mathbf{x}_{b}, \mathbf{x}_{c}$}
		\EndFunction

		\While {$t < t_{max}$}
		\State $\mathbf{x}^t_a$ $\gets$ \Call{Sitandwait}{$\mathbf{x}^t$}
		\State $\mathbf{x}^t_b,\mathbf{x}^t_c$ $\gets$ \Call{Aggressive}{$\mathbf{x}^t$}
		\State Get next position $\mathbf{x}^{t+1}$ via Eq. (\ref{eq.15}) to (\ref{eq.18}) 
		\EndWhile
		\State \Return $\mathbf{x}_{best}, y_{best}$
	\end{algorithmic}
\end{algorithm}

\section{Experimental results and discussions}

\begin{table}[htbp]
	\centering 
	\setlength{\tabcolsep}{0.6mm}{
		\caption{Unimodal test function}\label{tab.97}
		\renewcommand\arraystretch{2}
		\begin{tabular}{lccc}
			\toprule 
			Function      &      Dim        &      Range             & $F_{min}$\\
			\midrule 
			$F_1(x) =  {\textstyle \sum_{i=1}^{n}x^2_i} $        & 30 & $[-100, 100]$  & 0 \\
			$F_2(x) =  \sum_{i=1}^{n}|x_i|+\prod_{i=1}^{n}|x_i|$ & 30 & $[-10, 10]$    & 0 \\
			$F_3(x) =  \sum_{i=1}^{n}(\sum_{j=1}^{i}x_j)^2 $     & 30 & $[-100, 100]$  & 0 \\
			$F_4(x) =  max_i\{|x_i|,1\le i\le n\} $              & 30 & $[-100, 100]$  & 0 \\
			$F_5(x) =  \sum_{i=1}^{n-1}[100(x_{i+1} - x^2_i)^2+(x_i-1)^2] $    & 30 & $[-30, 30]$  & 0 \\
			$F_6(x) = \sum_{i=1}^{n}(\left \lfloor x_i+0.5 \right \rfloor^2 )$ & 30 & $[-100, 100]$  & 0 \\
			$F_7(x) = \sum_{i=1}^{n} ix_i^4+random[0,1)$                       & 30 & $[-1.28, 1.28]$  & 0\\
			\hline 
	\end{tabular}}
\end{table}

\begin{table}[]
	\centering 
	
	\caption{Summary of the CEC’17 Test Functions}\label{tab.17}
	\renewcommand\arraystretch{1.4}
	\setlength{\tabcolsep}{0.8mm}{
		\begin{tabular}{llll}
			\hline
			& No. & Functions $(\mathbb{F}_i)$                                                                                           &  $\mathbb{F}_{min}$    \\ \hline
			\multirow{2}{*}{\begin{tabular}[c]{@{}l@{}}Unimodal\\      Functions\end{tabular}}                 & 1   & Shifted and Rotated Bent Cigar   Function                                                            & 100  \\ 
			& 2   & Shifted and Rotated Zakharov   Function                                                              & 200  \\ \hline
			\multirow{7}{*}{\begin{tabular}[c]{@{}l@{}}Simple\\      Multimodal\\      Functions\end{tabular}} & 3   & Shifted and Rotated Rosenbrock’s   Function                                                          & 300  \\  
			& 4   & Shifted and Rotated Rastrigin’s   Function                                                           & 400  \\ 
			& 5   & \begin{tabular}[c]{@{}l@{}}Shifted and Rotated Expanded   Scaffer’s F6 \\ Function\end{tabular}      & 500  \\ 
			& 6   & \begin{tabular}[c]{@{}l@{}}Shifted and Rotated Lunacek   Bi\_Rastrigin \\ Function\end{tabular}      & 600  \\ 
			& 7   & \begin{tabular}[c]{@{}l@{}}Shifted and Rotated   Non-Continuous Rastrigin’s \\ Function\end{tabular} & 700  \\ 
			& 8   & Shifted and Rotated Levy   Function                                                                  & 800  \\ 
			& 9   & Shifted and Rotated Schwefel’s   Function                                                            & 900  \\ \hline
			\multirow{10}{*}{\begin{tabular}[c]{@{}l@{}}Hybrid\\      Functions\end{tabular}}                  & 10  & Hybrid Function 1 (N=3)                                                                              & 1000 \\ 
			& 11  & Hybrid Function 2 (N=3)                                                                              & 1100 \\  
			& 12  & Hybrid Function 3 (N=3)                                                                              & 1200 \\ 
			& 13  & Hybrid Function 4 (N=4)                                                                              & 1300 \\ 
			& 14  & Hybrid Function 5 (N=4)                                                                              & 1400 \\  
			& 15  & Hybrid Function 6 (N=4)                                                                              & 1500 \\  
			& 16  & Hybrid Function 6 (N=5)                                                                              & 1600 \\  
			& 17  & Hybrid Function 6 (N=5)                                                                              & 1700 \\  
			& 18  & Hybrid Function 6 (N=5)                                                                              & 1800 \\ 
			& 19  & Hybrid Function 6 (N=6)                                                                              & 1900 \\ \hline
			\multirow{10}{*}{\begin{tabular}[c]{@{}l@{}}Composition\\      Functions\end{tabular}}             & 20  & Composition Function 1 (N=3)                                                                         & 2000 \\  
			& 21  & Composition Function 2 (N=3)                                                                         & 2100 \\  
			& 22  & Composition Function 3 (N=4)                                                                         & 2200 \\  
			& 23  & Composition Function 4 (N=4)                                                                         & 2300 \\  
			& 24  & Composition Function 5 (N=5)                                                                         & 2400 \\  
			& 25  & Composition Function 6 (N=5)                                                                         & 2500 \\  
			& 26  & Composition Function 7 (N=6)                                                                         & 2600 \\  
			& 27  & Composition Function 8 (N=6)                                                                         & 2700 \\  
			& 28  & Composition Function 9 (N=3)                                                                         & 2800 \\  
			& 29  & Composition Function 10 (N=3)                                                                        & 2900 \\ \hline
	\end{tabular}}
\end{table}

\begin{figure*}[t]

		\centering
		
		\subfigure[F1 in CEC 1997]{\includegraphics[scale=0.25]{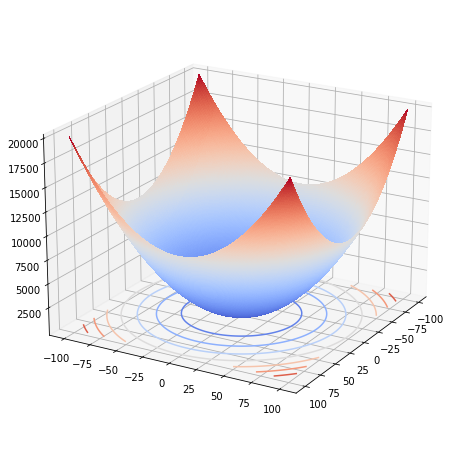}}
		\subfigure[F4 in CEC 1997]{\includegraphics[scale=0.25]{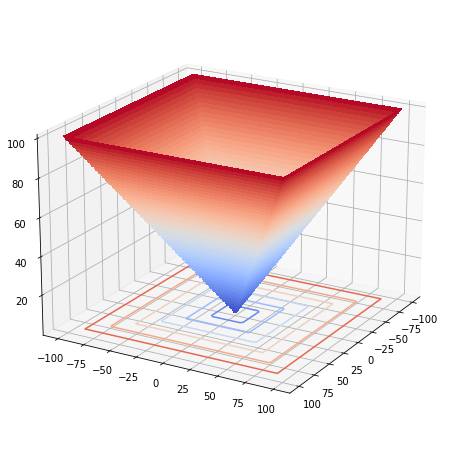}}
		\subfigure[$\mathbb{F}_6$ in CEC 2017]{\includegraphics[scale=0.25]{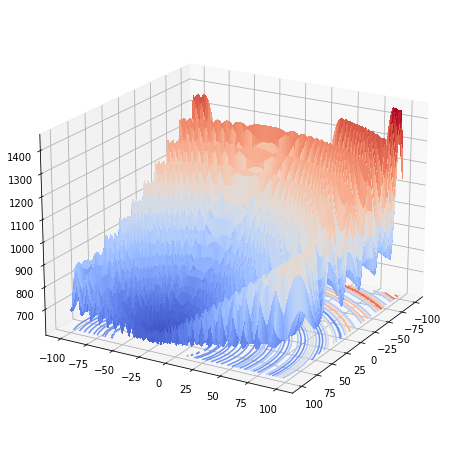}}
		\subfigure[$\mathbb{F}_8$ in CEC 2017]{\includegraphics[scale=0.25]{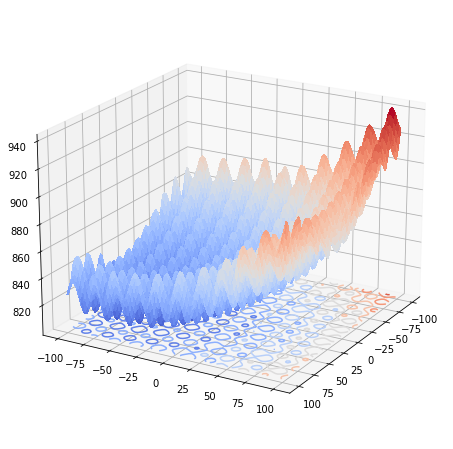}}
		\subfigure[$\mathbb{F}_{22}$ in CEC 2017]{\includegraphics[scale=0.25]{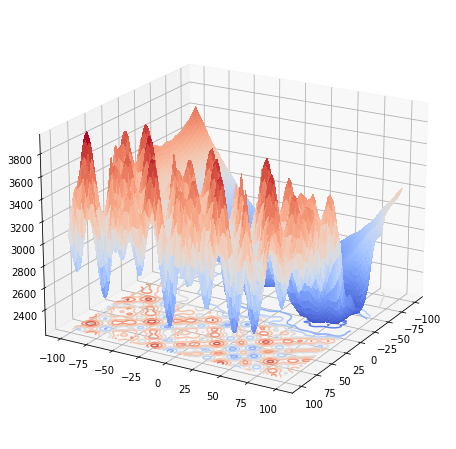}}
		\subfigure[$\mathbb{F}_{24}$ in CEC 2017]{\includegraphics[scale=0.25]{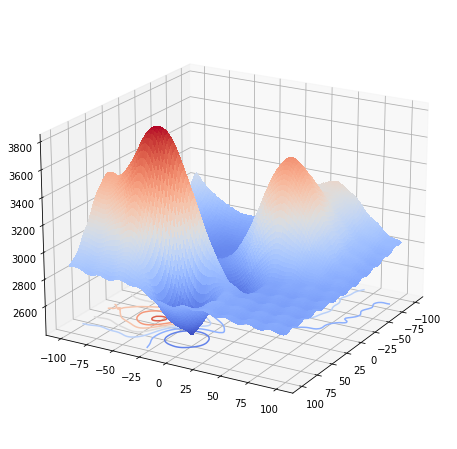}}
		\subfigure[$\mathbb{F}_{26}$ in CEC 2017]{\includegraphics[scale=0.25]{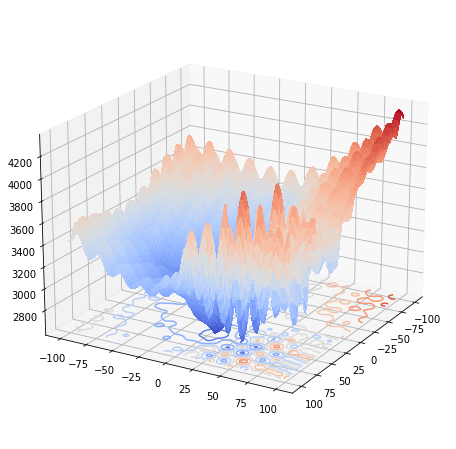}}
		\subfigure[$\mathbb{F}_{28}$ in CEC 2017]{\includegraphics[scale=0.25]{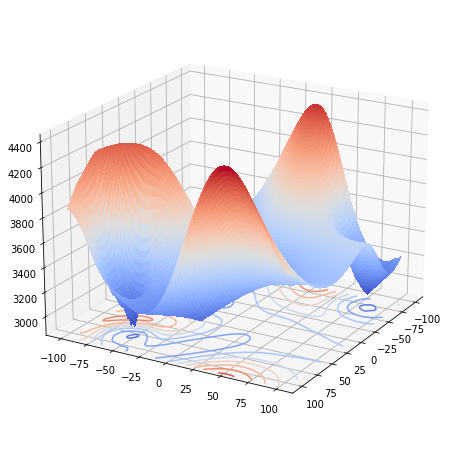}}

	\caption{(a) and (b) are Unimodal Functions, (c) and (d) are Simple Multimodal Functions, (e), (f), (g) as well as (h) are Composition Functions }\label{fig.ex1}
	
\end{figure*}

\begin{table*}[]
	\caption{Comparison Of Optimization Results Under The Unimodal Functions\\Dimension = 30, Maximum Iterations = 500}\label{tab.comp_1}

		\renewcommand\arraystretch{1.4}
		\setlength{\tabcolsep}{1.6mm}{
			\begin{tabular}{llllllllllllll}
				\hline
				F  &  & \multicolumn{2}{l}{ESOA} & \multicolumn{2}{l}{PSO\cite{ref.9}} & \multicolumn{2}{l}{GA\cite{ref.3}} & \multicolumn{2}{l}{DE\cite{ref.4}} & \multicolumn{2}{l}{GWO\cite{ref.11}} & \multicolumn{2}{l}{HHO\cite{ref.32}} \\
				&  & $ave$         & $std$        & $ave$        & std        & $ave$        & $std$       & $ave$        & $std$       & $ave$        & $std$        & $ave$        & $std$        \\ \hline
				$F_1$ &  & 0.00E+00    & 0.00E+00   & 1.89E+04   & 1.10E+04   & 8.87E+00   & 1.69E+01  & 7.94E-05   & 1.99E-05  & 8.85E-37   & 5.70E-37   & 1.09E-74   & 1.89E-74   \\
				$F_2$ &  & 4.92E-184   & 0.00E+00   & 1.12E+03   & 1.61E+02   & 1.76E+00   & 9.19E-01  & 3.25E-02   & 6.06E-03  & 1.64E-22   & 2.84E-22   & 9.03E-43   & 1.56E-42   \\
				$F_3$ &  & 0.00E+00    & 0.00E+00   & 3.72E+04   & 1.10E+04   & 1.21E+04   & 3.07E+03  & 1.88E+04   & 3.08E+03  & 2.54E-22   & 2.75E-22   & 2.58E-39   & 4.46E-39   \\
				$F_4$ &  & 3.53E-173   & 0.00E+00   & 2.98E+01   & 6.32E+00   & 2.80E+01   & 2.79E+00  & 2.73E+00   & 1.82E-01  & 5.16E-22   & 3.86E-22   & 2.58E-39   & 4.46E-39   \\
				$F_5$ &  & 2.81E+01    & 3.46E-01   & 1.01E+10   & 6.31E+09   & 2.55E+03   & 4.34E+03  & 1.75E+02   & 4.59E+01  & 1.24E-21   & 9.80E-22   & 4.96E-38   & 8.00E-38   \\
				$F_6$ &  & 5.20E+00    & 3.82E-01   & 2.04E+04   & 1.24E+04   & 6.25E-01   & 8.51E-01  & 1.21E-04   & 3.45E-05  & 1.14E-21   & 1.05E-21   & 5.00E-38   & 7.97E-38   \\
				$F_7$ &  & 2.26E-05    & 2.25E-05   & 7.64E+08   & 5.48E+08   & 3.05E+01   & 3.40E+01  & 1.79E-01   & 2.47E-02  & 4.00E-10   & 6.93E-10   & 4.75E-38   & 8.11E-38  \\ \hline
		\end{tabular}}

\end{table*}

\begin{table*}[]
	\caption{Comparison Of Optimization Results Under CEC17 Test Functions\\Dimension = 30, Maximum Iterations = 500}\label{tab.comp_2}

		\renewcommand\arraystretch{1.5}
		\setlength{\tabcolsep}{1.6mm}{
			\begin{tabular}{lllllllllllll}
				\hline
				$\mathbb{F}$  & ESOA     &          & PSO\cite{ref.9}      &          & GA\cite{ref.3}        &          & DE\cite{ref.4}        &          & GWO\cite{ref.11}      &          & HHO\cite{ref.32}      &          \\
				& $ave$         & $std$        & $ave$        & std        & $ave$        & $std$       & $ave$        & $std$       & $ave$        & $std$        & $ave$        & $std$        \\ \hline
				1  & 1.11E+09 & 2.07E+08 & 5.80E+11 & 1.41E+11 & 1.51E+07  & 2.10E+07 & 6.52E+06  & 3.91E+06 & 2.21E+10 & 1.52E+10 & 4.25E+11 & 1.32E+10 \\
				2  & N/A      & N/A      & N/A      & N/A      & N/A       & N/A      & N/A       & N/A      & N/A      & N/A      & N/A      & N/A      \\
				3  & 37289.32 & 6900.14  & 83594.42 & 24578.12 & 158232.44 & 49994.95 & 120278.68 & 8522.19  & 50388.30 & 9531.38  & 86299.93 & 5549.89  \\
				4  & 515.24   & 17.97    & 5635.36  & 3472.52  & 583.92    & 39.82    & 546.68    & 7.84     & 619.33   & 93.38    & 9697.10  & 1905.41  \\
				5  & 624.51   & 11.18    & 806.12   & 36.59    & 628.97    & 31.19    & 682.86    & 17.12    & 622.88   & 44.97    & 857.15   & 26.65    \\
				6  & 642.70   & 6.02     & 680.31   & 4.24     & 642.52    & 5.97     & 612.21    & 1.47     & 625.00   & 7.32     & 691.28   & 8.59     \\
				7  & 963.55   & 20.38    & 1976.48  & 473.69   & 1046.79   & 94.89    & 925.90    & 17.28    & 975.61   & 21.04    & 1355.05  & 16.84    \\
				8  & 942.24   & 14.78    & 1127.26  & 27.92    & 922.53    & 33.92    & 991.15    & 8.16     & 921.27   & 66.19    & 1105.98  & 20.14    \\
				9  & 3423.14  & 1023.40  & 8887.78  & 2054.96  & 2871.54   & 705.51   & 1068.04   & 68.50    & 3224.27  & 1127.23  & 6684.65  & 698.78   \\
				10 & 5016.83  & 231.31   & 6076.19  & 630.65   & 4393.26   & 659.06   & 7960.00   & 248.19   & 5755.16  & 1888.33  & 7759.34  & 253.55   \\
				11 & 1360.73  & 22.66    & 2034.61  & 340.33   & 8536.84   & 7784.18  & 1450.17   & 40.98    & 1809.70  & 827.89   & 9619.31  & 1658.95  \\
				12 & 3.66E+07 & 1.00E+07 & 1.98E+10 & 7.40E+09 & 1.08E+07  & 6.32E+06 & 3.99E+07  & 1.42E+07 & 9.10E+08 & 1.16E+09 & 4.28E+10 & 2.13E+10 \\
				13 & 3.73E+06 & 9.01E+05 & 3.51E+10 & 2.95E+10 & 1.89E+05  & 3.41E+05 & 1.16E+07  & 9.36E+06 & 3.77E+06 & 4.84E+06 & 1.86E+10 & 1.23E+10 \\
				14 & 4.35E+04 & 4.59E+04 & 7.38E+04 & 4.67E+04 & 1.32E+06  & 1.02E+06 & 1.73E+05  & 4.03E+04 & 9.29E+05 & 1.30E+06 & 6.45E+06 & 4.64E+06 \\
				15 & 2.20E+05 & 1.14E+05 & 1.05E+10 & 6.13E+09 & 1.60E+04  & 9.15E+03 & 9.86E+05  & 4.10E+05 & 2.19E+05 & 2.24E+05 & 9.20E+08 & 1.72E+09 \\
				16 & 2590.12  & 94.97    & 3767.87  & 573.77   & 3018.88   & 348.22   & 3160.53   & 142.81   & 2478.57  & 326.71   & 5353.46  & 924.34   \\
				17 & N/A      & N/A      & N/A      & N/A      & N/A       & N/A      & N/A       & N/A      & N/A      & N/A      & N/A      & N/A      \\
				18 & 1.88E+05 & 6.06E+04 & 4.03E+06 & 6.21E+06 & 4.40E+06  & 3.37E+06 & 2.00E+06  & 4.07E+05 & 1.33E+06 & 1.12E+06 & 2.65E+07 & 2.32E+07 \\
				19 & 1.21E+06 & 5.37E+05 & 3.97E+09 & 6.97E+09 & 2.82E+03  & 7.52E+02 & 1.90E+06  & 1.02E+06 & 1.69E+06 & 1.65E+06 & 1.07E+09 & 8.98E+08 \\
				20 & N/A      & N/A      & N/A      & N/A      & N/A       & N/A      & N/A       & N/A      & N/A      & N/A      & N/A      & N/A      \\
				21 & 2420.72  & 11.39    & 2574.28  & 66.54    & 2422.36   & 40.61    & 2483.75   & 10.18    & 2408.42  & 32.18    & 2674.48  & 22.91    \\
				22 & 2346.94  & 4.66     & 7280.93  & 783.05   & 3996.07   & 2106.12  & 3129.04   & 751.21   & 6665.97  & 1644.22  & 8699.15  & 1063.76  \\
				23 & 2811.31  & 13.06    & 3053.03  & 108.80   & 2802.26   & 26.61    & 2838.09   & 8.08     & 2809.62  & 64.69    & 3405.99  & 95.73    \\
				24 & 2974.72  & 19.61    & 3195.71  & 49.61    & 2986.88   & 33.57    & 3028.09   & 7.32     & 2992.27  & 53.38    & 3478.17  & 142.44   \\
				25 & 2929.65  & 16.14    & 5816.72  & 1718.64  & 2939.10   & 33.40    & 2891.38   & 3.17     & 2984.29  & 26.92    & 4302.69  & 433.31   \\
				26 & 3100.04  & 64.72    & 8426.19  & 739.31   & 5729.40   & 313.67   & 5483.57   & 140.53   & 4669.66  & 234.40   & 10614.99 & 1175.87  \\
				27 & 3236.51  & 11.52    & 3398.10  & 89.95    & 3250.51   & 17.04    & 3266.37   & 7.45     & 3241.82  & 9.68     & 3993.86  & 217.88   \\
				28 & 3310.59  & 11.00    & 6759.69  & 1123.67  & 3312.93   & 29.35    & 3292.70   & 9.67     & 3468.25  & 66.19    & 6214.76  & 303.99   \\
				29 & N/A      & N/A      & N/A      & N/A      & N/A       & N/A      & N/A       & N/A      & N/A      & N/A      & N/A      & N/A     \\ \hline
		\end{tabular}}

\end{table*}

\begin{figure*}[htbp]\centering

		\subfigure[F1 in CEC 1997]{\includegraphics[scale=0.22]{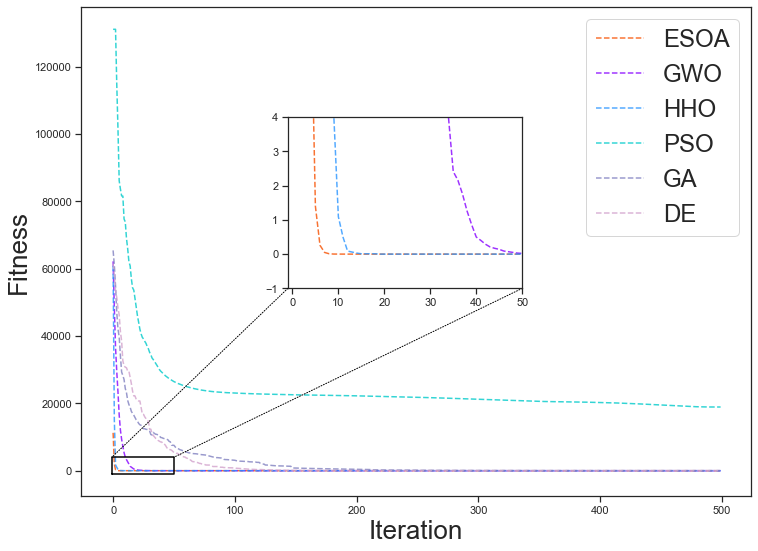}}
		\subfigure[F4 in CEC 1997]{\includegraphics[scale=0.22]{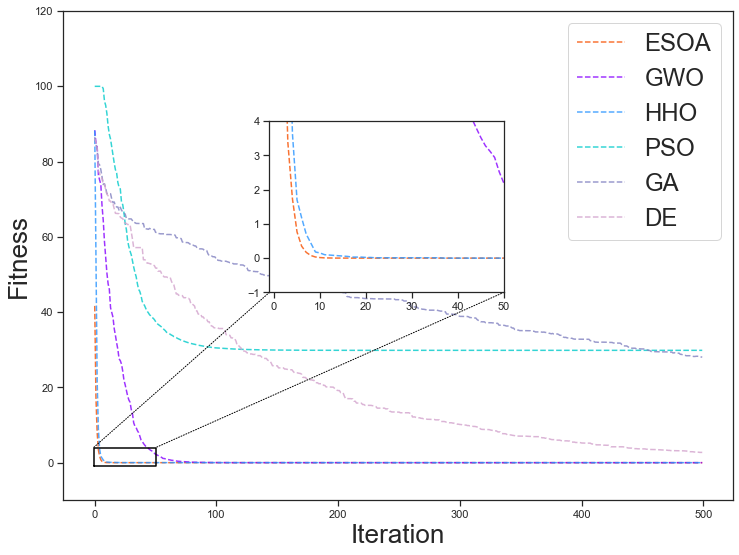}}
		\subfigure[F7 in CEC 1997]{\includegraphics[scale=0.22]{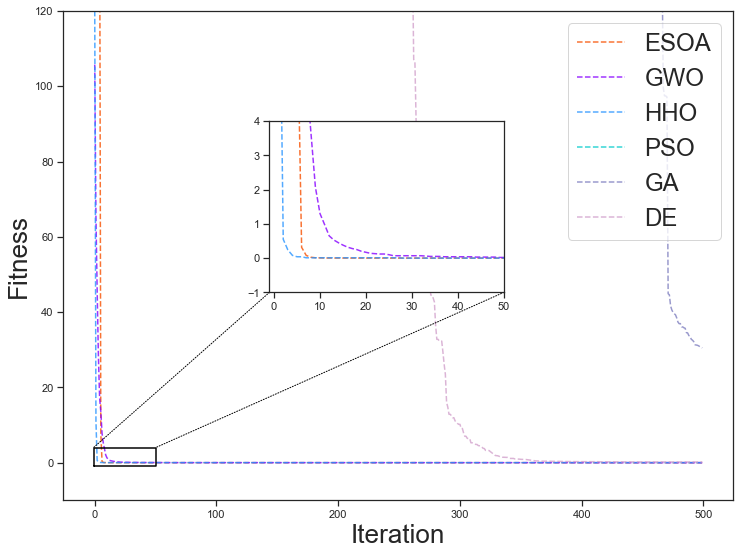}}
		\subfigure[$\mathbb{F}_4$ in CEC 2017]{\includegraphics[scale=0.22]{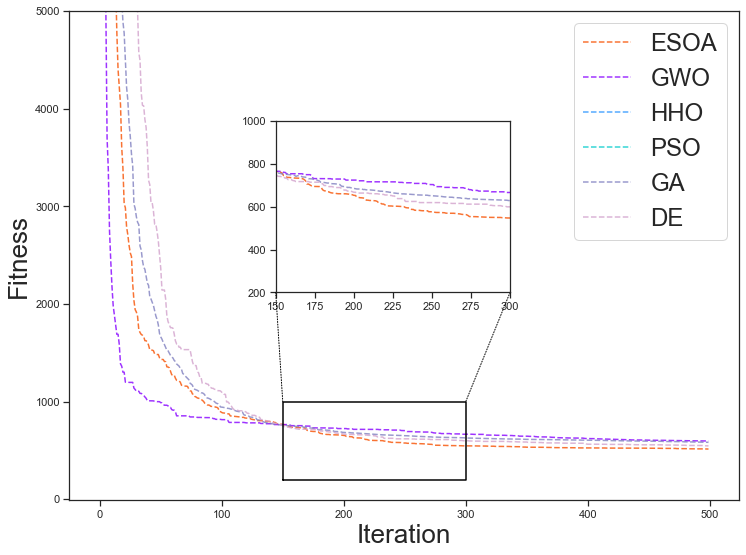}}
		\subfigure[$\mathbb{F}_{11}$ in CEC 2017]{\includegraphics[scale=0.22]{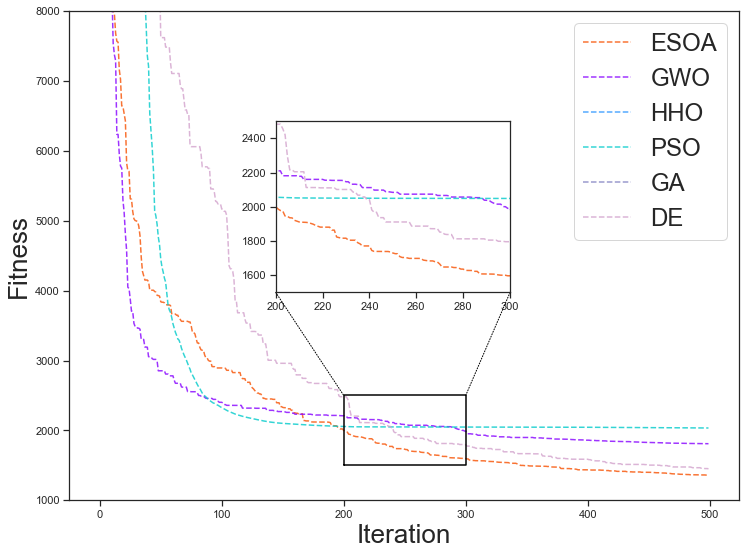}}
		\subfigure[$\mathbb{F}_{16}$ in CEC 2017]{\includegraphics[scale=0.22]{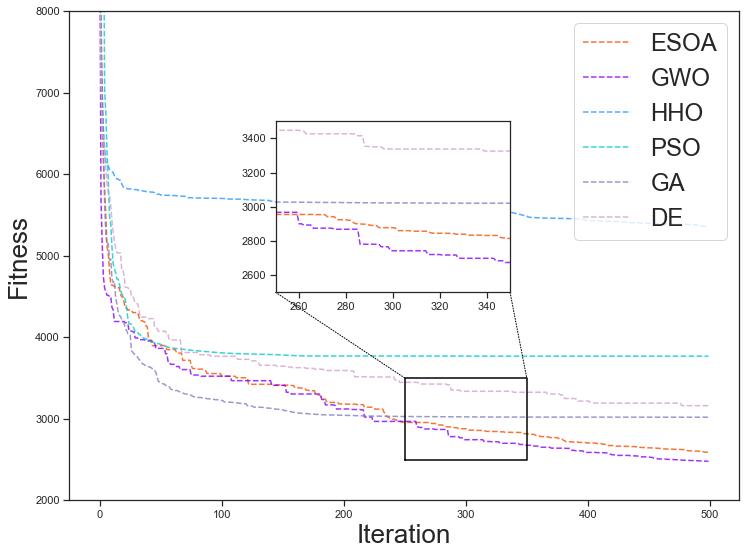}}
		\subfigure[$\mathbb{F}_{21}$ in CEC 2017]{\includegraphics[scale=0.22]{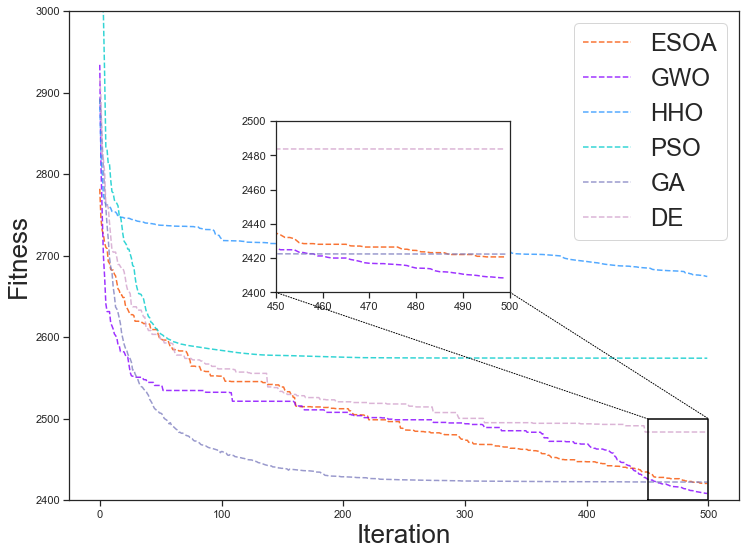}}
		\subfigure[$\mathbb{F}_{24}$ in CEC 2017]{\includegraphics[scale=0.22]{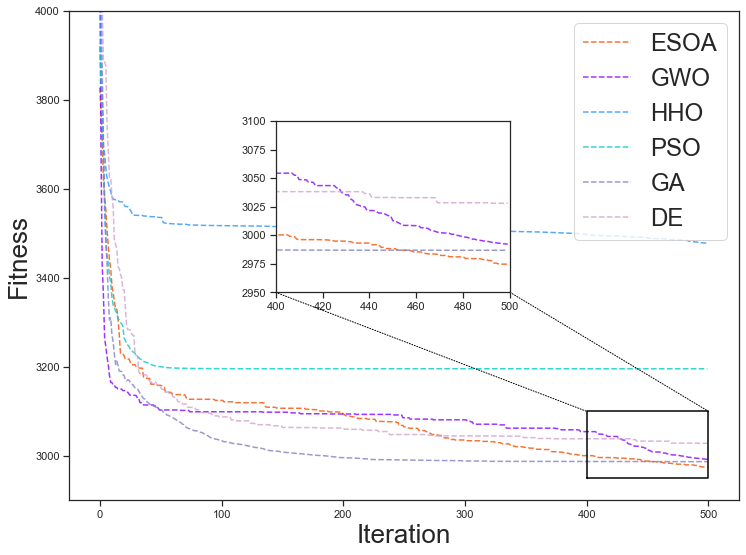}}
		\subfigure[$\mathbb{F}_{26}$ in CEC 2017]{\includegraphics[scale=0.22]{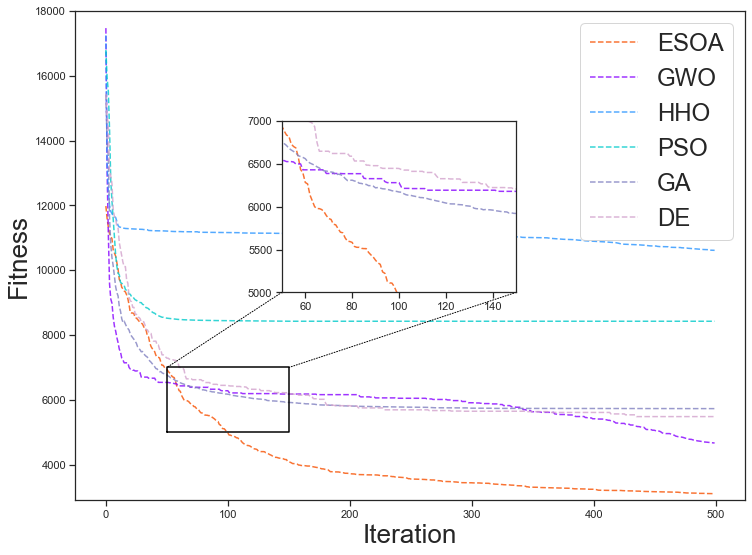}}

	\caption{(a), (b), (c) are Unimodal Functions, (d) is Simple Multimodal Functions, (e) and (f) are Hybrid Functions, (g), (h) as well as (i) are Composition Functions }\label{fig.ex2}
\end{figure*}

In this section, the quantified performance of the ESOA algorithm is retrieved via examining 36 optimization functions. The first 7 unimodal functions are typical benchmark optimization problems presented by \cite{ref.30} and the mathematical expressions, dimensions, the range of solution space as well as the best fitness are indicated in Table \ref{tab.97} . The final result and partial convergence curves are shown in Table \ref{tab.comp_1} as well as Fig. \ref{fig.ex2} respectively. The remaining 29 test functions introduced in \cite{ref.31} are constructed by summarizing valid features from other benchmark problems, such as cascade, rotation, shift as well as shuffle traps. The overview of these functions are shown in Table \ref{tab.17} and the comparison is indicated in Table \ref{tab.comp_2}. the whole experiments are in 30 dimensions, and the algorithms utilized are with 50 population sizes and 500 maximum iterations.

Researchers generally classify optimization test functions as Unimodal, Simple Multimodal, Hybrid Functions, and Composition Functions. As it is shown in Fig. \ref{fig.ex1}, 3D visualizations of several functions are indicated. As Unimodal Functions, (a) and (b) are $F_1$ as well as $F_4$ in CEC 1997, which only have one global minimum value without local optima. (c) and (d), the Simple Multimodal problems, retain numerous local optimal solution traps and multiple peaks to impede the exploration of global optimal search. Hybrid Multimodal functions are a series of problems adding up several different test functions with well-designed weights, and due to the dimension restriction, these are hard to reveal in the 3D graphics. (e), (f), (g) as well as (h) are Composition Functions,  the nonlinear combination of numerous test functions, extremely hard to optimize because of their local traps in various places and a variety of spikes to obstacle the algorithm's derivation.

ESOA was compared with three traditional algorithms (PSO \cite{ref.9}, GA \cite{ref.3}, DE \cite{ref.4}) as well as two novel methods (GWO \cite{ref.13}, HHO \cite{ref.32}) in the 37 benchmark functions. The numerical result from a maximum of 500 iterations is presented in Table \ref{tab.comp_1} and Table \ref{tab.comp_2}. The initial input of each algorithm is a random matrix with 30 dimensions and 50 populations in $[-100, 100]$. The specific variables $w$, $c_1$, and $c_2$ in PSO are 0.8, 0.5, and 0.5 while the mutation value is 0.001 in GA.

\subsection{Evaluation Of Exploitation Ability}
The Unimodal Function is utilized to evaluate the convergence speed and exploitation ability of the algorithms, as only one global optimum point is present. As presented in Table \ref{tab.comp_1}, ESOA demonstrates outstanding performace from $F_1$ to $F_4$. ESOA trails GWO and HHO in $F_5$ to $F_7$,  however, the result is considerably superior to PSO, GA, as well as DE. Therefore, the excellent exploitation of ESOA is evident here.

\subsection{Evaluation Of Exploration Ability($\mathbb{F}_3$-$\mathbb{F}_{19}$)}
In MultiModal Functions and Hybrid Functions, there are numerous local optimum positions for deceiving the algorithm into halting, and the optimization difficulty increases exponentially with rising dimensions, which are worth evaluating the exploration capability of an algorithm. The result of $\mathbb{F}_3$-$\mathbb{F}_{19}$ shown in Table \ref{tab.comp_2} provides powerful proof of the remarkable exploration ability of ESOA. In particular, ESOA presents superior performance to the other five algorithms from the average fitness on $\mathbb{F}_{18}$. Because of the Aggressive Strategy component of ESOA, it is capable of overcoming the interference of numerous local optimum points in the exploration of the global solution.

\subsection{Comprehensive Performance Assessment($\mathbb{F}_{20}$-$\mathbb{F}_{29}$)}
Composition Functions are a difficult type of test function that necessitates a great balance between exploitation and exploration of the algorithm under test. They are usually employed to undertake comprehensive evaluations of algorithms. The performance of each algorithms in $\mathbb{F}_{20}$-$\mathbb{F}_{29}$ is indicated in Table \ref{tab.comp_2}, the average fitness of ESOA in each test function is extremely competitive among the listed algorithms. Especially in $\mathbb{F}_{22}$, ESOA outperforms the other approaches and reaches 2346 average fitness while the second method DE only obtains 3129. In fact, the proposed algorithm ESOA possesses Sit-And-Wait Strategy for exploitation as well as an Aggressive Strategy for exploration. These features are regulated by a discriminant condition which is fundamental to the performance of the algorithm in such scenarios. 

\subsection{Algorithm Stability}
In general, the standard deviation of an algorithm's outcomes when it is repeatedly applied to a problem can reflect its stability. It can be seen that the standard deviation of the ESOA is at the top results in both tables in most situations, and much ahead of the second position in certain test functions. The stability of ESOA is hence proven.

\subsection{Analysis Of Convergence Behavior}
The partial convergence curves of each method are shown in Fig. \ref{fig.ex2}. In (a), (b), and (c), the Unimodal Functions, ESOA converges to near the global optimum in less than 10 iterations while PSO, GA as well as DE have yet to uncover the optimal path for fitness descent. The fast convergence in unimodal tasks allows ESOA to be applied to some online optimization problems. In (d), (e), and (f), the Multimodal and Hybrid Functions, after a period of searching the optimization results of ESOA will surpass almost all other algorithms in most cases and would continue to explore afterward. ESOA's effectiveness in Multimodal problems indicates that it has notable potential to be applied in general engineering applications. In (g), (h) as well as (i), the Composition problems, ESOA's search, and estimation mechanism allow for continuous optimization in most cases, and ultimately for excellent results. The performance in Composition Functions evidence of ESOA's applicability for use in complex engineering applications.

In conclusion, this section revealed ESOA's properties under various test functions. The Sit-And-Wait Strategy in ESOA allows the algorithm to perform fast descent on deterministic surfaces. ESOA's Aggressive Strategy ensures that the algorithm is extremely exploratory and does not readily slip into local optima. Therefore, ESOA has shown excellent outcomes in both exploration and exploitation.

\section{Typical Application}

In this section, ESOA is utilized in two practical engineering applications to demonstrate its competitive capabilities in optimization constrain problems. In order to simplify the computational procedure, the penalty function is adopted to integrate the inequality constraints into the objective function \cite{ref.33}. The specific form is as below,

\begin{equation}
	\hat{f} (\mathbf{x}) = f (\mathbf{x}) + \phi \sum_{j=1}^{p}g^2_j(\mathbf{x}) sgn(g_j(\mathbf{x})) \label{eq.plenaty},
\end{equation}

\begin{equation}
	sgn(g_j(\mathbf{x})) = \left\{\begin{matrix} 
		1, \qquad if g_j(\mathbf{x})>0 \\  
		0, \qquad if g_j(\mathbf{x})\le0. 
	\end{matrix}\right. 
\end{equation}
where $\hat{f} (\mathbf{x})$ is the transformed objective function and $\phi$ mean the penalty parameter while $f (\mathbf{x})$ and $g_j(\mathbf{x})$ present the origin objective function as well as the inequality constrains respectively, $j\in[1,2,...,p]$ and $p$ is the number of constrains. $sgn(g_j(\mathbf{x}))$ is used to determine whether the independent variable violates a constrains.

\subsection{Himmelblau’s Nonlinear Optimization Problem}
Himmelblau introduced a nonlinear optimization problem as one of the famous benchmark problems for meta-heuristic algorithms \cite{ref.34}. The problem is describled as below,

\begin{table}[htbp]
	\caption{The Statistical Result Of Various Algorithms For Himmelblau Problem}\label{tab.h}
	\renewcommand\arraystretch{1.5}
	\setlength{\tabcolsep}{1.5mm}{
		\begin{tabular}{llllll}
			\hline
			Algorithm &  & Best         & Worst        & Ave          & Std      \\ \hline
			ESOA      &  & -30665.18896 & -30062.49696 & -30524.47712 & 136.7795737 \\
			PSO\cite{ref.9}       &  & -30663.21135 & -29194.42152 & -29959.04964 & 399.3063815 \\
			GA\cite{ref.3}        &  & -30347.35433 & -28638.48333 & -29440.23438 & 437.0557374 \\
			DE \cite{ref.4}        &  & -30655.14151 & -29421.63637 & -30395.93485 & 273.0411217 \\
			GWO\cite{ref.11}       &  & -30663.47493 & -30503.84528 & -30643.9799  & 27.38941651 \\
			HHO\cite{ref.32}       &  & -30495.629   & -28093.40301 & -29229.05025 & 511.6730218 \\ \hline
	\end{tabular}}
\end{table}

\begin{equation*}
	\begin{split}
		&Minimize \,\, f(\mathbf{x}) = 5.3578547 x^2_3 + 0.8356891 x_1 x_5 \\
		& \qquad \qquad\qquad\qquad + 37.293239 x_1 - 40792.141\\ \\
		&s.t.\quad  \left\{\begin{array}{lc}
			g_1(\mathbf{x}) = 85.334407 + 0.0056858 x_2 x_5 \\
			\qquad \qquad + 0.0006262 x_1 x_4 - 0.0022053 x_3 x_5\\\\
			g_2(\mathbf{x}) = 80.51249 + 0.0071317 x_2 x_5 \\
			\qquad \qquad + 0.0029955 x_1 x_2 - 0.0021813 x^2_3\\ \\
			g_3(\mathbf{x}) = 9.300961 + 0.0047026 x_3 x_5 \\
			\qquad \qquad + 0.0012547 x_1 x_3 - 0.0019085 x_3 x_4\\\\
			0 \le g_1(\mathbf{x}) \le 92\\ \\
			90 \le g_2(\mathbf{x}) \le 11\\ \\
			20 \le g_3(\mathbf{x}) \le 25\\  \\
			78 \le x_1\le102 \\ \\
			33 \le x_2\le45 \\ \\
			27 \le x_3\le45 \\ \\
			27 \le x_4\le45 \\ \\
			27 \le x_5\le45. \\ 
		\end{array}\right.
	\end{split}
\end{equation*}

For this problem, the $\phi$ in Eq. (\ref{eq.plenaty}) is set to $10^{100}$, the number of search agents used by each algorithm is set to 10, and the maximum number of iterations is set to 500. 

The statistic result from 30 trials for each algorithm is shown in Table \ref{tab.h}. ESOA is the best performing algorithm in terms of optimal results, with the best result reaching $-30665.18896$. The second best was achieved by PSO at $-30663.21135$. The standard deviation represents the algorithm's stability, and ESOA, although ranking second , outperforms PSO, GA, DE, as well as HHO. The experimental results demonstrate the engineering feasibility of the proposed method.

\subsection{Tension/Compression Spring Design}

\begin{figure}[htbp]\centering
	
	\includegraphics[scale=0.065]{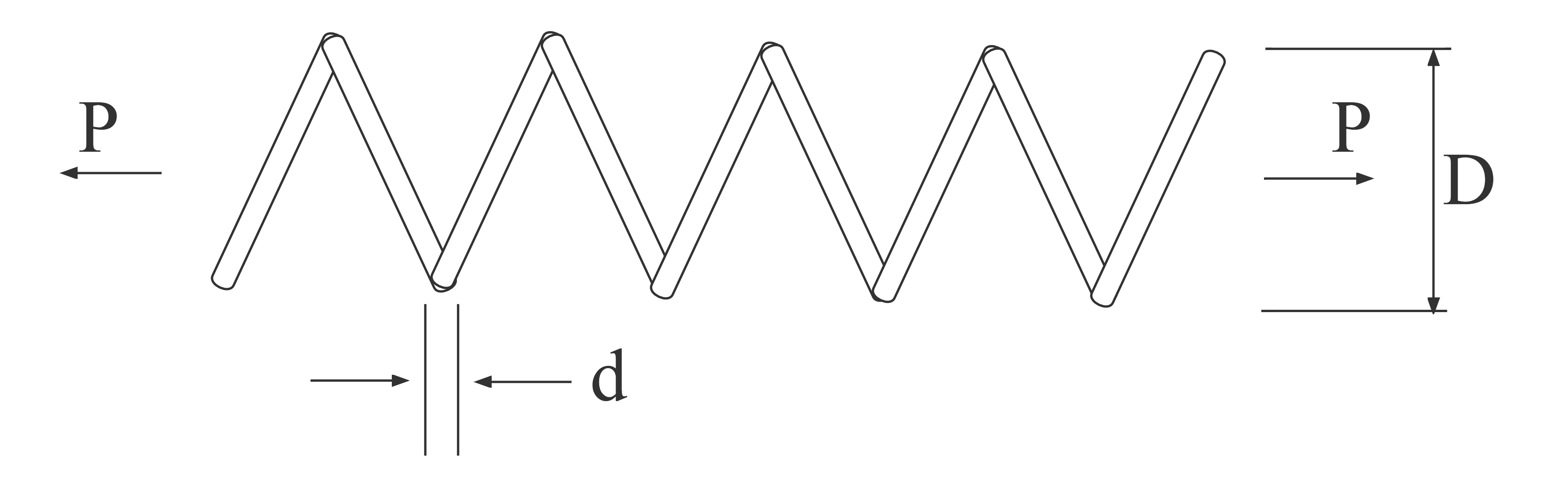}
	
	\caption{Tension/compression string design problem.}\label{fig.string}
\end{figure}

The string design problem is described by Arora and Belegundu for minimizing spring weight under the constraints of minimum deflection, shear stress, and surge frequency \cite{ref.35, ref.36}. Fig. \ref{fig.string} illustrates the details of the problem, $P$ is the number of active coils, $d$ means the wire diameter while $D$ represents the mean coil diameter.

The mathematical modeling is as below,

\begin{equation*}
	\begin{split}
		&Minimize \,\, f(\mathbf{x}) = (x_3 + 2) x_2 x^2_1\\
		&s.t.\quad  \left\{\begin{array}{lc}
			g_1(\mathbf{x}) = 1 - \frac{x^3_2x_3}{71785 x^4_1} \le  0\\\\
			g_2(\mathbf{x}) = \frac{4 x^2_2 - x_1 x_2}{12566(x_2 x^3_1-x^4_1)} + \frac{1}{5108 x^2_1} - 1  \le  0\\ \\
			g_3(\mathbf{x}) = 1 - \frac{140.45 x_1}{x^2_2 x_3} \le  0\\\\
			g_4(\mathbf{x}) = \frac{x_1 + x_2}{1.5} - 1 \le  0.\\
		\end{array}\right.
	\end{split}
\end{equation*}

\begin{table}[htbp]
	\caption{The Statistical Result Of Various Algorithms For Spring Problem}\label{tab.s}
	\renewcommand\arraystretch{1.5}
	\setlength{\tabcolsep}{2.8mm}{
		\begin{tabular}{llllll}
			\hline
			Algorithm &  & Best     & Worst    & Ave      & Std      \\\hline
			ESOA      &  & 0.012743 & 0.013023 & 0.012855 & 7.34E-05 \\
			PSO\cite{ref.9}       &  & 0.012719 & 39721.55 & 1324.065 & 7130.235 \\
			GA\cite{ref.3}        &  & 0.013089 & 11658.45 & 464.3166 & 2102.06  \\
			DE\cite{ref.4}        &  & 0.012715 & 0.02629  & 0.015176 & 0.002537 \\
			GWO\cite{ref.11}       &  & 0.012695 & 0.014352 & 0.012892 & 0.000307 \\
			HHO\cite{ref.32}       &  & 0.012727 & 0.016021 & 0.014132 & 0.001111 \\\hline
	\end{tabular}}
\end{table}

For this problem, the $\phi$ in Eq. (\ref{eq.plenaty}) is set to $10^{5}$, the number of search agents used by each algorithm is again set to 10, and the maximum number of iterations is set to 500.

Table \ref{tab.s} indicates the statistic result from 30 independent runs for the six algorithms. The average fitness of ESOA achieves the best result at $0.12855$, whilst GWO achieves the second best result at $0.012892$. The standard deviation of ESOA outperforms other algorithms, which demonstrates ESOA's exceptional stability. The proposed algorithm's (ESOA) effectiveness and robustness are hence verified.

\section{Conclusion}

This paper introduced a novel meta-heuristic algorithm, the Egret Swarm Optimization Algorithm, which mimics two egret species' (Great Egret and Snowy Egret) typical hunting behavior. ESOA consists of three essential components: Snowy Egret's Sit-And-Wait Strategy, Great Egret's Aggressive Strategy as well as a Discriminant Condition. The performance of ESOA was then compared with 5 other state-of-the-art methods (PSO, GA, DE, GWO, and HHO) on 36 test functions, including Unimodal, Multimodal, Hybrid, and Composition Functions. The results of which demonstrate ESOA's exploitation, exploration, comprehensive performance, stability as well as convergence behavior. In addition, two practical engineering problem instances demonstrate the excellent performance and robustness of ESOA to typical optimization applications.

To accommodate more applications and optimization scenarios, other mathematical forms of Sit-And-Wait Strategy, Aggressive Strategy, and Discriminant Condition in ESOA are currently under development.

\end{document}